\documentclass[10pt,twocolumn,letterpaper]{article}

\usepackage{iccv}
\usepackage{times}
\usepackage{epsfig}
\usepackage{graphicx}
\usepackage{amsmath}
\usepackage{amssymb}
\usepackage{xcolor}
\usepackage{pifont}
\usepackage[normalem]{ulem} 
\usepackage{booktabs}
\usepackage{tabularx}
\usepackage{float}
\usepackage{amsfonts}
\usepackage{authblk}
\usepackage{multirow}
\usepackage{float}
\usepackage{capt-of}

\newcommand{\blue}[1]{{\color{blue}#1}}

\usepackage[pagebackref=true,breaklinks=true,letterpaper=true,colorlinks,bookmarks=false]{hyperref}

\iccvfinalcopy 

\def\iccvPaperID{****} 

\ificcvfinal\fi

\begin{document}

\definecolor{custom_yellow}{HTML}{FFC107} 
\definecolor{custom_blue}{HTML}{1E88E5}

\title{The Change You Want to See\break(Now in 3D)\vspace{-10pt}}

\author{Ragav Sachdeva}
\author{Andrew Zisserman}
\affil{Visual Geometry Group, Dept.\ of Engineering Science, University of Oxford}

\twocolumn[{
    \vspace{-30pt}
    \renewcommand\twocolumn[1][]{#1}
    \maketitle
    \centering
    \vspace{-30pt}
    \includegraphics[width=\textwidth]{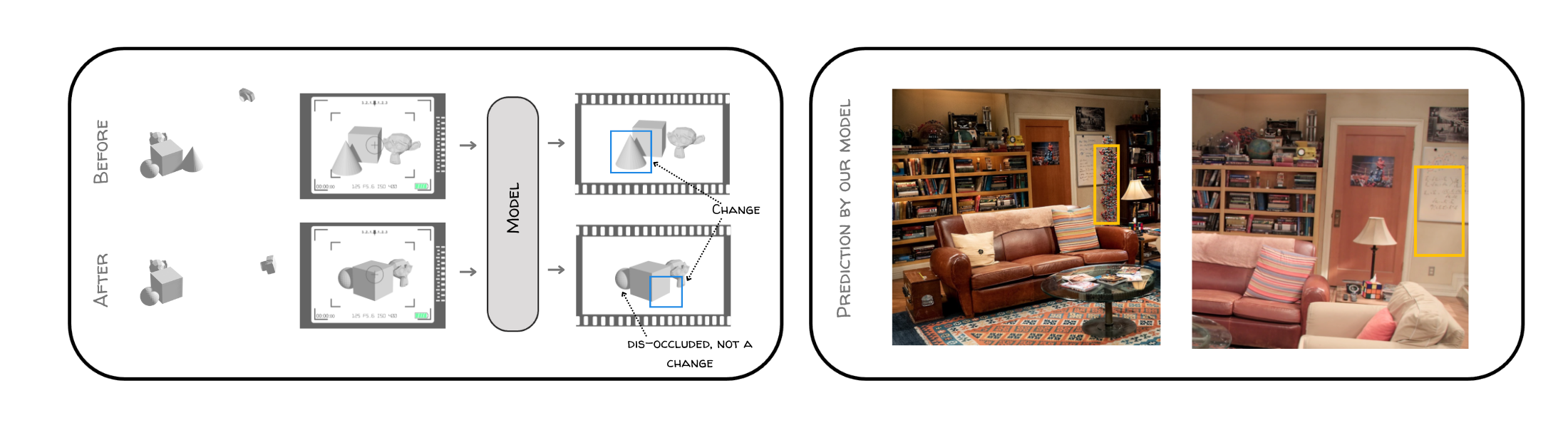}
    \captionof{figure}{\textbf{On the left:} An illustration of the problem we aim to address in this work. Given an image pair of a 3D scene, with a significant shift in camera pose, can we detect what has changed? \textbf{On the right:} Top prediction by our model on images from the popular TV show \textit{The Big Bang Theory} depicting the missing DNA-model.}
    \label{fig:teaser}
    \vspace{20pt}}]
    
\ificcvfinal\thispagestyle{empty}\fi

\begin{abstract}
The goal of this paper is to detect what has changed, if anything, between two ``in the wild" images of the same 3D scene acquired from different camera positions and at different temporal instances. The open-set nature of this problem, occlusions/dis-occlusions due to the shift in viewpoint, and the lack of suitable training datasets, presents substantial challenges in devising a solution.

To address this problem, we contribute a change detection model that is trained entirely on \textbf{synthetic data} and is \textbf{class-agnostic}, yet it is \textbf{performant out-of-the-box on real world images} without requiring fine-tuning. Our solution entails a ``register and difference" approach that leverages self-supervised frozen embeddings and feature differences, which allows the model to generalise to a wide variety of scenes and domains. The model is able to \textbf{operate directly on two RGB images}, without requiring access to ground truth camera intrinsics, extrinsics, depth maps, point clouds, or additional before-after images. Finally, we collect and \textbf{release a new evaluation dataset} consisting of real-world image pairs with human-annotated differences and demonstrate the efficacy of our method. The code, datasets and pre-trained model can be found at: \blue{\url{https://github.com/ragavsachdeva/CYWS-3D}}
\end{abstract}


\section{Introduction}

\noindent From the way leaves rustle in the wind to the shifting patterns of clouds in the sky, our world is in a constant state of flux. Yet detecting and localising changes in complex 3D scenes remains a challenging task for computer vision. Imagine being able to identify the changes between two images of the same scene captured at separate moments in time and from different viewpoints, as shown in Figure~\ref{fig:teaser}. This is the challenge we aim to address in this work. With applications in fields such as robotics, facility monitoring, forensics and augmented reality, our work has the potential to unlock new possibilities for understanding and interacting with our dynamic world.

We formulate the problem we study as follows: Given a pair of 2D images of a 3D scene, captured with a significant shift in camera position and at different temporal instances, we wish to localise the changes between them, if any. In particular, we wish to capture \textit{everything} that is physically different \textbf{in the regions that are visible in both the images} while disregarding areas that appear or disappear from view due to the shift in camera pose or occlusion. This includes objects that may have been added or removed from the scene, and text or decorations that may have been added to an object, but we wish to ignore photometric differences such as a lighting change.
Under this setting, differentiating true changes from occlusions or dis-occlusions is intrinsically ill-posed using 2D images alone. In other words, answering the question ``Is this object missing in the other image, hidden behind another object, or simply out of frame?"
fundamentally requires the 3D shape of the scene, which is not directly available to us apriori. Furthermore, it is not possible to compute the shape of the given 3D scene using two-view stereo methods as they rely on corresponding points in the two images which are inherently non-existent in case of changes such as missing objects.
Consequently, in the absence of the scene's ground truth geometry, it is theoretically infeasible to reason about the relative position, scale and shape of the objects in the scene, and how or where they might appear when observed from a different viewpoint (see Figure~\ref{fig:teaser}). This, coupled with the lack of large-scale real-world datasets that include image pairs of changing scenes captured from significantly different viewpoints, makes devising a solution to this general two-view change detection problem very challenging.

Nevertheless, our objective is to detect changes in ``in the wild" real world images with minimal constraints and operate on RGB images only, without assuming access to ground truth geometry, depth, camera parameters, camera poses etc.
Our solution is to build on the insight that once the two views are \textit{registered}, it is relatively easy to determine what has changed. We thus proceed in two stages: (1) \textit{register} by warping the spatial features from one image to the other, and this fundamentally needs to be in 3D; (2) \textit{determine the differences} by training a detector on top of these registered spatial feature maps in order to identify the significant changes, and ignore ``nuisance" variables such as changes in lighting. Briefly, we first use an off-the-shelf pre-trained visual backbone to extract spatial feature maps for an RGB image pair. We then lift these 2D feature maps to 3D by making use of state-of-the-art monocular depth estimation models, followed by a differentiable feature registration module (DFRM) to align and render the features maps back to 2D in the other view. Finally, we utilise a simple detection head to process these features and output the changes. To overcome the issue of lack of real-world training data, we train our model exclusively on synthetic data with controlled 3D changes. In order to permit sim2real, we keep the visual backbone frozen throughout training, and decode \textit{difference} of registered features. 

Since the notion of change necessitates a pair of corresponding regions with some differences, our formulation requires model predictions in \textit{both} the images. For instance, if a car is present in one image but missing in the other, we expect the model to put bounding boxes around both -- where the car is, and where the car ``should have been". Furthermore, since the changes in ``in the wild" images are customarily \textbf{open-set}, our model is designed to be \textbf{class-agnostic}. We demonstrate that a model trained in this fashion zero-shot generalises to a wide variety of datasets including 2D scenes, 3D scenes, synthetic and real-world images.

In the following we provide an overview of the existing literature (Sec.~\ref{sec:related_works}), details of the proposed method (Sec.~\ref{sec:method}), experiments and results (Sec.~\ref{sec:experiments}), and finally some concluding remarks (Sec.~\ref{sec:conclusion}). We will release the code, datasets and trained model. 

\section{Related Work}
\label{sec:related_works}
\noindent The problem of exploring visual changes has been studied in several different flavours previously. In this section we loosely group these works into two categories, 2D and 3D, and describe how our problem setting relates/differs from them.\newline

\noindent\textbf{2D (-ish):} A typical scenario for the change detection problem is one where we have a pair of before-after RGB images, where the camera is either fixed i.e. the images are related by an identity transformation (e.g.\ images from surveillance cameras), there is a planar-scene assumption (e.g.\ bird's eye view or satellite images), or in the general case there is limited shift in viewpoint (e.g. street scenes looking at distant objects/buildings), and the model is expected to identify the changes between them. Most of the existing works in the change detection literature belong to this category.

\cite{robustchange, captiondiff, multiclevr, kimcap} tackle the change captioning problem where the goal is to describe the changes in an image pair in natural language. These methods mainly evaluate their approach on the STD~\cite{spotthediff} (images from fixed video surveillance camera), or CLEVR-based change datasets~\cite{robustchange, multiclevr, kimcap} (synthetic images of 3D objects of primitive shapes). Since the problem these works address is change captioning, their approaches do not deal with precise localisation of changed regions.

\begin{figure*}[t]
    \centering
    \includegraphics[width=\linewidth]{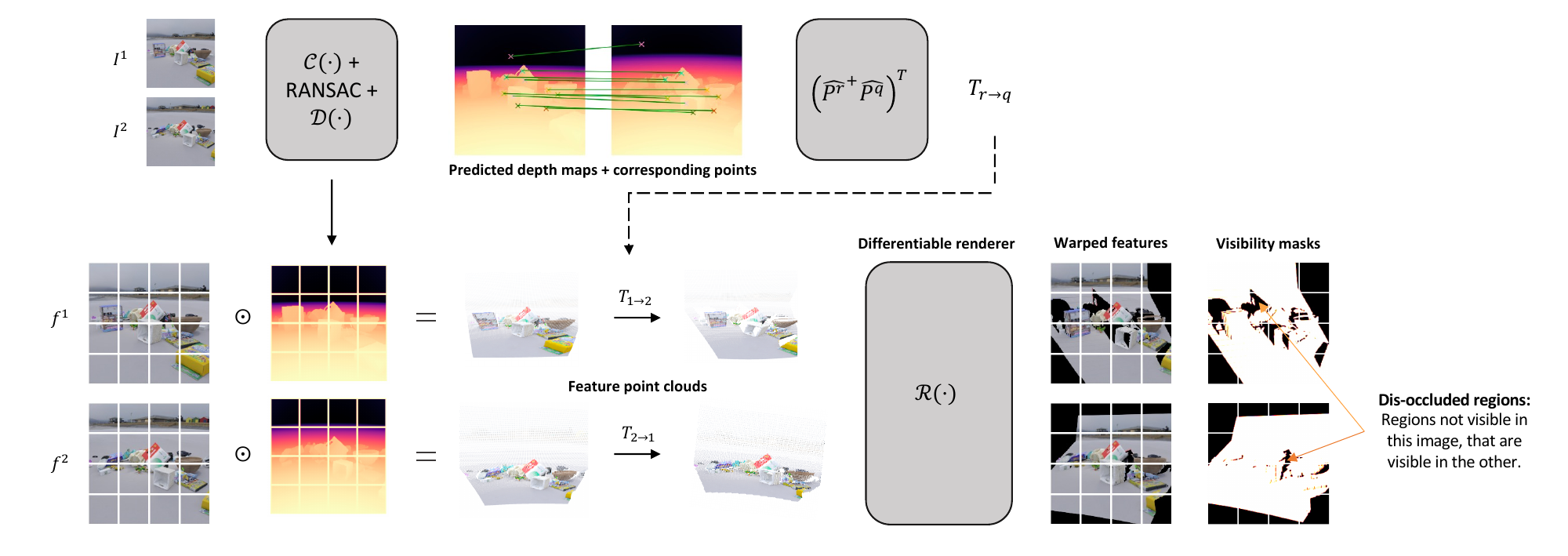}
    \vspace*{-7mm}
    \caption{\textbf{DFRM:} Given two images, we first obtain corresponding points and depth maps which are used to estimate the 3D transformation $T_{r\rightarrow q}$. Then given a feature map for each image along with the depth map, we create a point cloud of $c$-dimensional feature vectors, warp it using the estimated transformation and render it to 2D grid. Notice the black regions in the rendered grid. These regions are dis-occluded and contain $c$-dimensional $\boldsymbol{0}$ vectors. We obtain a soft visibility mask to counteract the effect of dis-occluded regions when computing the difference.}
    \label{fig:dfrm}
\end{figure*}

\cite{pcd, Stent-RSS-16, weakly, streetscenehei} tackle the change localisation problem, specifically for street-view images where the goal is to produce segmentation masks for changed regions. These methods mainly evaluate their approach on TSUNAMI~\cite{pcd}  (panaromic, street-view images), VL-CMU~\cite{Stent-RSS-16} (images of urban scenes with macroscopic changes) and PCSD~\cite{weakly} (panoramic, street-view images). While these datasets do not technically have the planar-scene assumption or a fixed camera, the scenes are of distant buildings and there is no ``peeking behind" objects due to a shift in camera pose. Since pixel-wise change annotation is expensive, these datasets often suffer from \textit{non-comprehensively labelled test sets, limited to a set classes}.

Recently, \cite{Sachdeva23} proposed a \textit{class-agnostic} method that tackles the change detection problem for arbitrary images that are related by a homography transformation, as a bounding-box based detection problem. They evaluate their approach on STD~\cite{virat}, Kubric-Change~\cite{Sachdeva23} (3D objects resting on a 2D plane, camera is not fixed but the images are captured from bird's eye view) and others.

Similar to the methods above, we also tackle the change detection problem in a pair of RGB images. However, unlike previous methods our setting involves \textbf{general two-view images of 3D scenes} (we do not assume fixed camera or planar scenes and in our case there is a significant shift in camera pose) and we particularly focus on making our model work on \textbf{open-set}, real-world images. The problem formulation closest to ours is that of~\cite{Sachdeva23} in that ~\cite{Sachdeva23} also train a class-agnostic model that produces bounding boxes around changed regions in both the images except they train and evaluate on 2D scenes.\newline

\noindent \textbf{3D:} The change detection problem has also been studied explicitly for 3D data (e.g.\ multiview before-after images, 3D scans etc.). \cite{multiview_captioning} tackle the change captioning problem in a 3D setting by assuming multi-view images are available both before and after the change. \cite{multimodality_captioning} further propose an end-to-end framework for describing scene changes from various input modalities, namely, RGB images, depth images, and point cloud data. Recently, \cite{qiuwacv} proposed a task to explicitly localise changes in the form of 3D bounding boxes from two point clouds and describe detailed scene changes \textit{for a fixed classes of objects}. Unlike these works, we only assume access to a single-view image for both before and after scenes and \textbf{do not operate on explicit 3D data} like point clouds. Since RGB images are more readily available, it allows our method to be easily applied to real-world scenes. The closest setting to our is of \cite{doi2022detecting} who tackle the change detection problem for general two-view images, except their model is trained and tested on the same synthetic setting with 4 fixed classes\footnote{\label{not_released}Not available publicly at the time of writing.}.

\section{Method}
\label{sec:method}

\subsection{Overview}
\noindent Given an image pair of a 3D scene, captured with a significant shift in camera viewpoints, our goal is to localise the changes between them in the form of bounding box predictions for \textit{each} image. In particular, we only wish to capture changes in the regions that are visible in both the images while disregarding areas that appear or disappear from view due to the shift in camera pose. 

\begin{figure*}[ht]
    \centering
    \includegraphics[width=\linewidth]{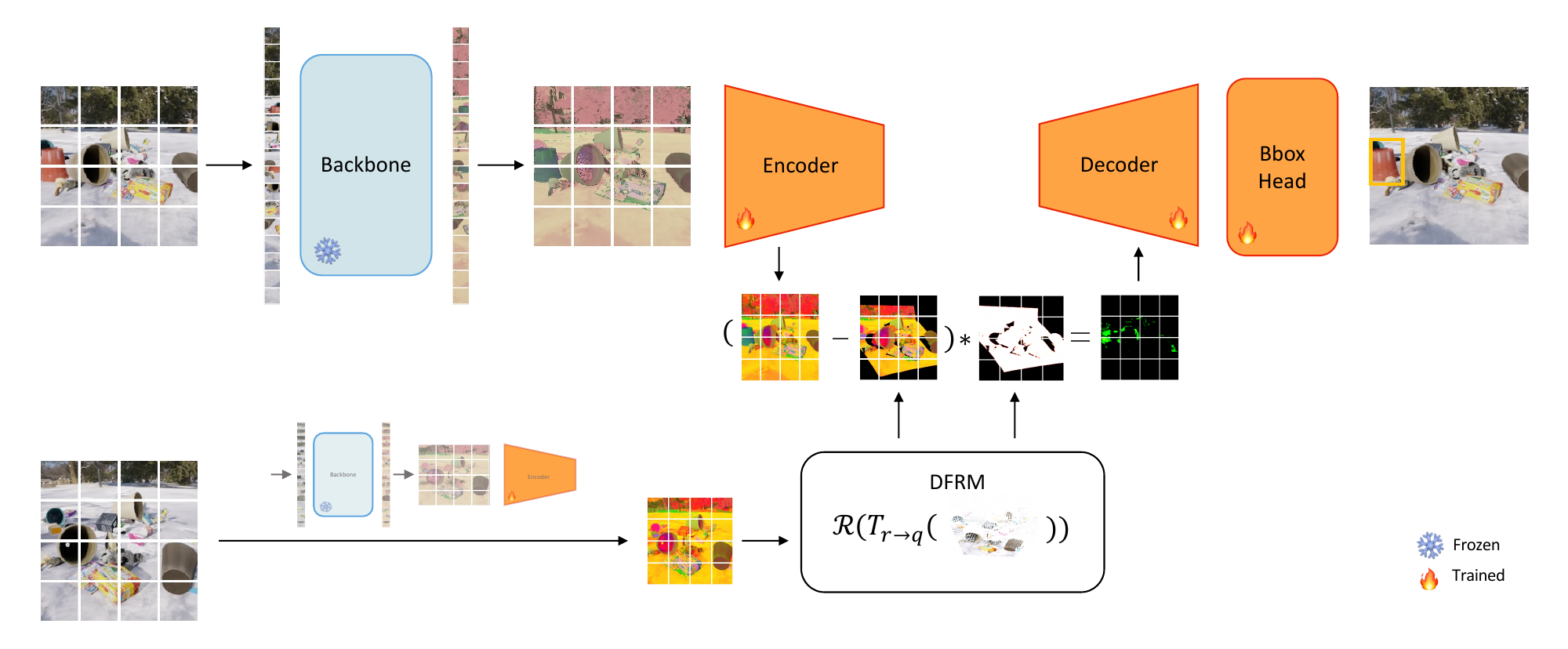}
    \caption{\textbf{Architecture:}  Given two images, we first extract dense spatial feature maps using a pre-trained visual backbone. Following this, a CNN-based encoder is used to extract visual descriptors at multiple spatial resolutions. We then use a differentiable feature registration module (DFRM) to warp features from one image to another (such that they are registered), and take their difference (only one resolution is shown for the purpose of visualisation, however this operation is performed at multiple resolutions). Finally, these difference of feature maps are processed by a CNN-based decoder followed by a bounding box detection head. For brevity, we only show the prediction for one of the images, however, the pipeline is symmetric for the other. Please see Sec.~\ref{sec:method} for more details.}
    \label{fig:model_arch}
\end{figure*}

Our approach, which is overviewed in Figure~\ref{fig:dfrm},  begins by extracting dense spatial image descriptors from each image using a pre-trained transformer-based visual backbone, which are then processed to obtain feature maps at multiple spatial resolutions using a U-Net~\cite{unet} style encoder. Next, in order to reason between changes and occlusions/dis-occlusions, we need 3D information. Inspired by~\cite{synsin}, our DFRM ``lifts" the features to 3D, and differentiably \textit{registers} and renders them. After obtaining registered features, we compute their \textit{difference} in order to identify what has changed. Finally, we decode the difference of registered features using a U-Net style decoder, followed by a bounding box prediction head. We next describe the architecture, which is illustrated in Figure~\ref{fig:model_arch}.

\subsection{Architecture}
\noindent \textbf{Backbone:} Given two images $I^1 \in \mathbb{R}^{3\times H\times W}$ and $I^2 \in \mathbb{R}^{3\times H\times W}$,
we first encode $I^1, I^2$ using a pre-trained frozen visual backbone, represented by $\Phi_{B}(\cdot)$, to obtain two sets of feature descriptors per image represented by $f^1_s, f^1_d = \Phi_{B}(I_1)$ and $f^2_s, f^2_d = \Phi_{B}(I_2)$. In practise, $\Phi_{B}(\cdot)$ is the ViT-B/8 \cite{vit} architecture, where $f_s$ and $f_d$ represent shallow and deep features.\newline

\noindent \textbf{U-Net Encoder:} We process $f^1_d, f^2_d$ using a U-Net style encoder, represented by $\Phi_{E}(\cdot)$, to obtain $g^1_n=\Phi_{E}(f^1_d)_n$ and $g^2_n=\Phi_{E}(f^2_d)_n$ after each downsampling block $n$, resulting in a set of multi-resolution feature maps $G^1 = \{f^1_d, g^1_1, g^1_2, g^1_3, g^1_4\}$ and $G^2 = \{f^2_d, g^2_1, g^2_2, g^2_3, g^2_4\}$, for image $I^1$ and $I^2$ respectively.\newline

\noindent \textbf{Feature Registration and Difference:} Given $G^1, G^2$, we use a differentiable feature registration module (described in Sec.~\ref{sec:dfrm}) to obtain a \textit{warp} of feature maps at each spatial resolution such that the original features of one image are registered with the warped features of the other image. We then compute their element-wise difference and mask out occluded/dis-occluded regions. Specifically, for feature maps at spatial resolution $i$, we obtain the feature map $H^1_i = v_{2\rightarrow 1}(G^1_i - \tau_{2\rightarrow 1}(G^2_i))$ and $H^2_i = v_{1\rightarrow 2}(G^2_i - \tau_{1\rightarrow 2}(G^1_i))$, for $I^1, I^2$ respectively, where $\tau_{r\rightarrow q}$ represents the 3D feature warp operator from image $I^r$ to $I^q$, and $v_{r\rightarrow q}$ represents the \textit{soft} visibility mask.\newline

\noindent\textbf{U-Net Decoder:} Following this, we decode the set of feature maps $H^1$ and $H^2$ using a U-Net decoder modulated with scSE blocks \cite{scse}, represented by $\Phi_D(\cdot)$ to produce feature maps $k^1$ and $k^2$ respectively.\newline

\noindent\textbf{Bbox Head:} Finally, feature maps $[f^1_s \mathbin\Vert k^1]$ and $[f^2_s \mathbin\Vert k^2]$, where $[\ \mathbin\Vert\ ]$ is the concatenation operation (along channel dimension), are fed into a CenterNet head~\cite{centernet}, which minimises the detection loss function as described in~\cite{centernet}, to produce bounding boxes around changed regions in both the images. The motivation for concatenating shallow features $f_s$ with $k$ is that it serves as the final skip connection before the prediction head, which is consistent with the typical U-Net style model, and additionally shallow features are known to capture more \textit{positional} information, as opposed to deep features which capture more \textit{semantic} information~\cite{talidekel}, and therefore can help with localisation.

\subsection{Differentiable Feature Registration Module (DFRM)}
\label{sec:dfrm}
\noindent Given images $I^r, I^q \in \mathbb{R}^{3\times H\times W}$, along with their feature maps $f^r, f^q \in \mathbb{R}^{c\times h\times w}$ respectively, we use the following three-step process, represented by $\tau_{r\rightarrow q}(\cdot)$, to warp $f^r$ such that $f^q$ and $\tau_{r\rightarrow q}(f^r)$ are registered. See Figure~\ref{fig:dfrm} to visually conceptualise all the moving parts.\newline 

\noindent \textbf{Step 1: Estimate a 3D linear transformation.} In order to register the two feature maps, we must estimate a 3D transformation between the two images. To do so, first we obtain a set of $n$ corresponding points $P^r, P^q \in \mathbb{R}^{n\times 2}$ (in normalised coordinates) in image $I^r, I^q$ respectively, using a correspondence extractor represented by $\mathcal{C}(\cdot)$. Following this, we estimate the depth maps $D^r, D^q \in \mathbb{R}^{3\times H\times W}$ using a monocular depth estimator represented by $\mathcal{D}(\cdot)$. Finally, to estimate the transformation, we first back-project each point $(x_j, y_j) \in P^r, P^q$ as,

\begin{equation}
    \hat{P_j} = \begin{bmatrix}d_jx_j & d_jy_j & d_j & 1 \end{bmatrix}^T
\end{equation}

\noindent where $d_j$ is the depth value at point $P_j$, resulting in two sparse 3D point clouds $\hat{P^r}, \hat{P^q} \in \mathbb{R}^{n\times 4}$ in homogenous coordinates. Following this, we estimate a transformation matrix $T_{r\rightarrow q} \in \mathbb{R}^{4\times 4}$, such that $T_{r\rightarrow q}\hat{P^r_j} \approx \hat{P^q_j}$, using the following closed-form solution:

\begin{equation}
    T_{r\rightarrow q} = \left(\hat{P^r}^{+}\hat{P^q}\right)^{T}
\end{equation}

\noindent where $\hat{P^r}^{+} \in \mathbb{R}^{4\times n}$ is the Moore-Penrose inverse of $\hat{P^r}$.\newline

\noindent \textbf{Step 2: Lift the features to 3D and warp.} For each $c$-dimensional feature vector in $f^r \in \mathbb{R}^{c\times h\times w}$, we project its normalised 2D grid coordinates $(x_j, y_j)$ to a point $(\begin{smallmatrix}x_j^{'}/k, & y_j^{'}/k, & z_j^{'}/k\end{smallmatrix})$ in 3D, where

\begin{equation}
    \begin{bmatrix}x_j^{'} & y_j^{'} & z_j^{'} & k\end{bmatrix}^T = T_{r\rightarrow q} \begin{bmatrix}d_jx_j & d_jy_j & d_j & 1\end{bmatrix}^T
\end{equation}

\noindent where $d_j$ is the estimated depth value of this point, resulting in a 3D point cloud of $f^r$ feature vectors that are aligned with $f^q$. \newline

\noindent \textbf{Step 3: Differentiable feature rendering.} Given the 3D point cloud, we render it to the 2D grid using a differentiable renderer $\mathcal{R}(\cdot)$. However, instead of rendering RGB colours, we render $c$-dimensional feature vectors for each point. For this, we employ a differentiable point cloud renderer, as in~\cite{synsin}. The advantages of this differentiable point cloud renderer are two fold: (1) it solves the ``small neighbourhood" problem, wherein each feature-point projects to only one or a few pixels in the rendered view, by splatting the points to a disk of controllable size, and (2) the ``hard z-buffer" problem, wherein each rendered pixel is only affected by the nearest point in the z-buffer, by accumulating the effects of $K$ nearest points. Both of these allow for better gradient propagation during training. In addition to obtaining the rendered features, we also obtain a visibility mask $v_{r\rightarrow q}$ to deal with occluded/dis-occluded regions (see Figure~\ref{fig:dfrm}). This is obtained by setting the 2D coordinates of the rendered points to $1$ on a grid initialised with $0$s. However, due to the splatting behaviour of the renderer, the visibility mask obtained is \textit{soft} and not binary. \newline

\noindent \textbf{Alternate registration strategies:} Since DFRM is training-parameter free, during inference $\tau_{r\rightarrow q}(\cdot)$ may utilise alternate registration strategies. For instance, if ground truth depth is known for each image, it can directly replace $D^r, D^q$ obtained using the monocular depth estimator $\mathcal{D}(\cdot)$. In addition to ground truth depth, if the camera intrinsics and extrinsics are known, points can be directly back-projected and warped using relative camera poses without needing to estimate the transformation $T_{r\rightarrow q}$. Furthermore, if only $D^r$ is available, Perspective-n-Point methods~\cite{Hartley04c} can be used to warp features $f^r$ onto $f^q$.

On the other hand, if it is known apriori that the images are of planar scenes, depth is not needed and each 2D grid coordinate can be warped using $T_{r\rightarrow q}\in \mathbb{R}^{3\times 3}$ which can either be supplied or estimated using standard homography estimation methods. If the scene consists of multiple planes (e.g.\ floor and wall), it may also be possible to obtain desired results using a multi-grid~\cite{multigrid} or multi-plane homography estimation~\cite{Hartley04c} to register the images.

\subsection{Details and discussion}
\noindent The ViT-B/8~\cite{vit} backbone $\phi_B(\cdot)$ is initialised with DINO model weights~\cite{dino_vit}. DINO features are known to encode powerful high level semantic information at fine spatial granularity and recent works have shown that scalar product of DINO features can be used to compute high quality semantic correspondences~\cite{talidekel}. Orthogonally to computing correspondences, we utilise these powerful DINO features to compute changes. To prevent the model from overfitting to synthetic data and corrupting the quality of DINO features, we keep the backbone frozen during training. In order to allow sim2real, our decoder only ever operates on the {\em difference} of features, which is negatively proportional to the scalar product i.e.\ the higher the scalar product the smaller the difference, and therefore we only consider the similarity of features (or the lack of) rather than the features themselves (whether synthetic or real). Following~\cite{talidekel}, we increase the resolution of model features by changing the stride of the patch extraction (from 8 to 4) and adjusting the positional encoding appropriately via interpolation, allowing us to operate on more granular spatial features. The shallow features and deep features $f_s, f_d$ are extracted from the \textit{keys} of the Multiheaded Self-Attention layers of the third and the last block respectively, an insight also from~\cite{talidekel}.

For correspondence extractor $\mathcal{C}(\cdot)$, we experimented with~\cite{cotr, loftr} but found SuperGlue~\cite{superglue} to work the best at generating high quality dense correspondences and at high inference speed. After extracting 2D correspondences from SuperGlue, we filter out outliers using RANSAC. Finally, for the monocular depth estimator $\mathcal{D}(\cdot)$, we tried~\cite{midas, omnidata} but found the recently released ZoeDepth~\cite{zoe} to consistently produce better results.

\begin{figure*}[ht]
    \vspace{-30pt}
    \centering
    \includegraphics[width=\linewidth]{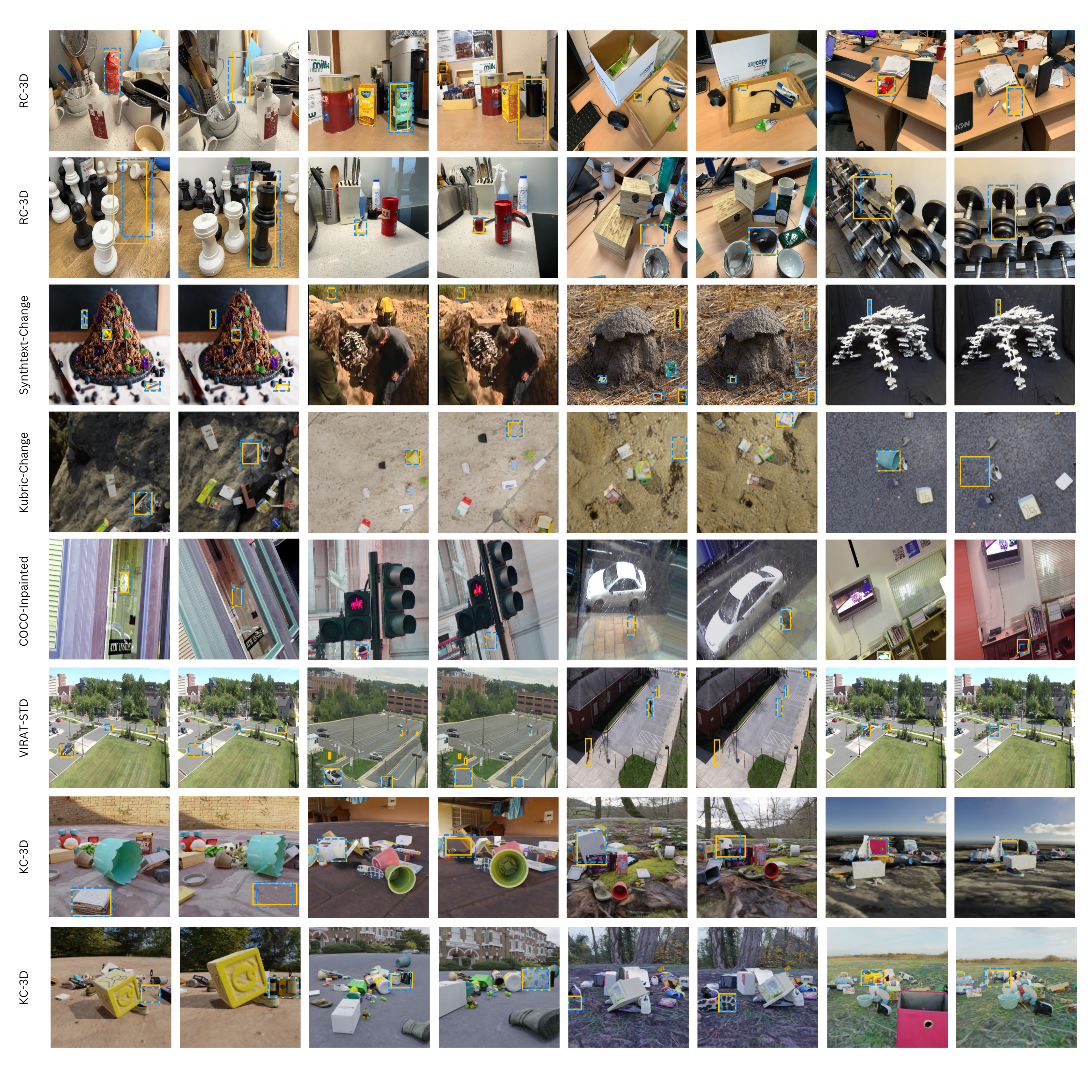}
    \caption{\textbf{Qualitative results:} We show the bounding box \textcolor{custom_yellow}{predictions in yellow (solid)} of our model on all the test sets, along with the \textcolor{custom_blue}{ground truth in blue (dashed)}.}
    \label{fig:predictions}
\end{figure*}

\begin{figure*}[ht]
    \vspace{-10pt}
    \centering
    \includegraphics[width=\linewidth]{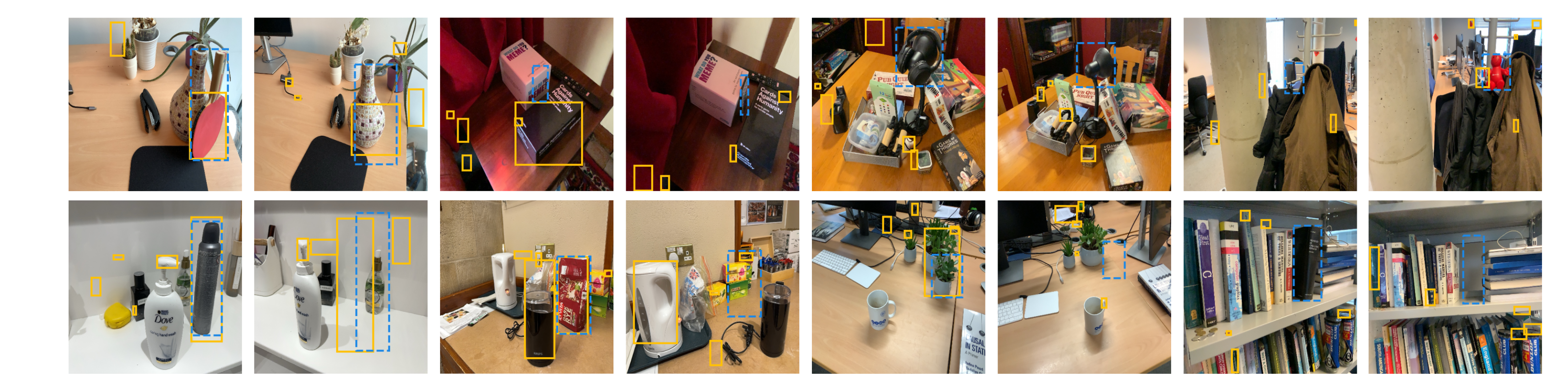}
    \caption{\textbf{Failure cases:} Here we show some RC-3D images where our model failed to localise the correct changed region. We classify an output as a \textit{failure} if the top-5 most confidence bounding boxes do not contain the ground truth changed region. \textcolor{custom_yellow}{Predictions are in yellow (solid)}, \textcolor{custom_blue}{ground truth is in blue (dashed)}.}
    \label{fig:failure}
\end{figure*}

\section{Experiments}
\label{sec:experiments}
\noindent This section describes the data we used to train our model, evaluation benchmarks and baselines, and various implementation details. Please see Table~\ref{tab:datasets} for an overview of the datasets used and Figure~\ref{fig:predictions} for some example images.

\subsection{Training datasets}
\label{sec:training_datasets}
\noindent Due to the lack of existing large-scale real-world datasets for the change detection problem, as formulated in this work, we fall-back to training our model entirely on synthetic data. Specifically, we train our model jointly on the following two datasets:\newline

\noindent \textbf{KC-3D:} Similar to \cite{Sachdeva23}, we make use of the Kubric dataset generator~\cite{kubric} to curate 86407 image pairs (of which 4548 images pairs are for validation) of 3D scenes with controlled changes. The scenes consist of randomly selected set of 3D objects spawned at random locations on a randomly textured plane, where we iteratively remove the objects and capture ``before" and ``after" image pairs. However, unlike~\cite{Sachdeva23} where they capture birds-eye view images only, we capture images from a wide variety of camera poses inside a cylindrical space around the objects. In addition, we also capture the depth maps, camera intrinsics and extrinsics for each image to supply to DFRM during training. \newline

\noindent \textbf{COCO-Inpainted:} While KC-3D captures the underlying challenges of our task in terms of viewpoint shift quite well, it lacks diversity in terms of kinds and sizes of objects. Therefore, we additionally utilise the recently introduced COCO-Inpainted dataset~\cite{Sachdeva23} for training. While this is a 2D dataset, where the image pairs are perturbed by an affine transformation, it helps the model learn to predict changes of various kinds and sizes.\newline

\subsection{Testing datasets}
\label{sec:testing_datasets}
\noindent To test the performance of our model, we evaluate it on the following test sets.\newline

\noindent \textbf{KC-3D:} Following the same pipeline as before, we curate an additional 4548 image pairs of 3D scenes from Kubric for testing purposes (making the total number of image pairs in KC-3D to be 90955). \newline

\noindent \textbf{RC-3D:} To quantify our model's capacity to generalise to real-world images, we manually collect and label a small-scale test set consisting of 100 images pairs, capturing a diverse set of common objects found in everyday places like office, kitchen, lounge etc. The images were captured using a handheld Apple iPad Pro (4th Gen) and Apple iPhone 14 Pro, which come with a built-in LiDAR giving us aligned RGB-D images. \newline

\noindent \textbf{Cyws Test Sets:} Since the problem we are tackling subsumes the change detection problem in planar scenes, we also evaluate our model on 2D test datasets proposed in~\cite{Sachdeva23}, namely COCO-Inpainted, VIRAT-STD, Kubric-Change, and Synthtext-Change.

\begin{table}[H]
\centering
\resizebox{\columnwidth}{!}{%
\begin{tabular}{l|llllll}
\hline
Test set       & COCO-Inpainted & Synthtext-Change & VIRAT-STD & Kubric-Change      & KC-3D & RC-3D \\ \hline
type           & inpainting     & text        & surveillance      & sim                & sim   & general 2-view  \\
change & synthetic & synthetic & real & synthetic & synthetic & real \\
geometry       & affine         & identity         & identity  & 2D-ish (bird's eye) & 3D    & 3D    \\
\# train images & 60000+           & -                & -         & -                  & 81859  & -     \\
\# test images  & 4408           & 5000             & 1000      & 1605               & 4548  & 100   \\ \hline
\end{tabular}%
}\newline
\caption{\textbf{Datasets.} KC-3D and RC-3D are ours. Others are from~\cite{Sachdeva23}.}
\label{tab:datasets}
\end{table}

\begin{table*}[ht]
\centering
\resizebox{\textwidth}{!}{%
\begin{tabular}{cccccccccccccclcl}
\hline
\multicolumn{2}{c}{\multirow{2}{*}{}} &
  \multicolumn{1}{c|}{\multirow{2}{*}{test dataset}} &
  \multicolumn{5}{c|}{COCO-Inpainted} &
  \multicolumn{1}{c|}{\multirow{2}{*}{VIRAT-STD}} &
  \multicolumn{1}{c|}{\multirow{2}{*}{Synthtext-Change}} &
  \multicolumn{1}{c|}{\multirow{2}{*}{Kubric-Change}} &
  \multicolumn{2}{c|}{\multirow{2}{*}{KC-3D}} &
  \multicolumn{4}{c}{\multirow{2}{*}{RC-3D}} \\
\multicolumn{2}{c}{} &
  \multicolumn{1}{c|}{} &
  small &
  medium &
  large &
  \multicolumn{2}{c|}{all} &
  \multicolumn{1}{c|}{} &
  \multicolumn{1}{c|}{} &
  \multicolumn{1}{c|}{} &
  \multicolumn{2}{c|}{} &
  \multicolumn{4}{c}{} \\ \hline
 &
   &
  \multicolumn{1}{c|}{depth} &
  \multicolumn{5}{c|}{Const.} &
  \multicolumn{1}{c|}{Const.} &
  \multicolumn{1}{c|}{Const.} &
  \multicolumn{1}{c|}{Const.} &
  \multicolumn{2}{c|}{GT} &
  \multicolumn{2}{c|}{GT} &
  \multicolumn{2}{c}{Est.} \\ \hline
\multicolumn{2}{c}{} &
  \multicolumn{1}{c|}{registration} &
  \multicolumn{3}{c|}{GT} &
  \multicolumn{1}{c|}{GT} &
  \multicolumn{1}{c|}{Est.} &
  \multicolumn{1}{c|}{Id.} &
  \multicolumn{1}{c|}{Id.} &
  \multicolumn{1}{c|}{Est.} &
  \multicolumn{1}{c|}{GT} &
  \multicolumn{1}{c|}{Est.} &
  \multicolumn{4}{c}{Est.} \\ \hline
method &
  \multicolumn{2}{c}{training data} &
  \multicolumn{14}{c}{(Const. = Constant, Id. = identity, Est. = Estimated, GT = Ground Truth)} \\ \hline
cyws &
  \multicolumn{2}{c}{coco-inpainted} &
  \textbf{0.46} &
  \textbf{0.79} &
  \textbf{0.85} &
  \multicolumn{2}{c}{\textbf{0.63}} &
  \textbf{0.65} &
  \textbf{0.89} &
  0.76 &
  \multicolumn{2}{c}{0.13} &
  \multicolumn{4}{c}{0.12} \\
cyws &
  \multicolumn{2}{c}{coco-inpainted + KC-3D} &
  0.41 &
  0.73 &
  0.78 &
  \multicolumn{2}{c}{0.57} &
  0.54 &
  0.87 &
  0.76 &
  \multicolumn{2}{c}{\textbf{0.87}} &
  \multicolumn{4}{c}{0.14} \\
ours &
  \multicolumn{2}{c}{coco-inpainted} &
  0.34 &
  0.69 &
  0.76 &
  0.52 &
  0.51 &
  0.46 &
  0.85 &
  \textbf{0.84} &
  0.14 &
  0.10 &
  \multicolumn{2}{c}{0.35} &
  \multicolumn{2}{c}{0.27} \\
ours &
  \multicolumn{2}{c}{KC-3D} &
  0 &
  0.03 &
  0.06 &
  0.02 &
  0.02 &
  0.01 &
  0.01 &
  0.23 &
  0.83 &
  0.69 &
  \multicolumn{2}{c}{0.19} &
  \multicolumn{2}{c}{0.19} \\
ours &
  \multicolumn{2}{c}{coco-inpainted + KC-3D} &
  0.36 &
  0.72 &
  0.77 &
  0.53 &
  0.52 &
  0.49 &
  0.84 &
  \textbf{0.84} &
  0.82 &
  0.68 &
  \multicolumn{2}{c}{\textbf{0.50}} &
  \multicolumn{2}{c}{0.41} \\ \hline
\end{tabular}%
}\newline
\caption{\textbf{Results:} We report the AP of cyws~\cite{Sachdeva23} and our model on test sets described in Sec~\ref{sec:testing_datasets}.}
\label{tab:results}
\end{table*}

\subsection{Baseline and metrics}
\noindent To the best of our knowledge, no prior works have tackled the change detection problem in a \textit{general two-view and class-agnostic} setting like us which makes it difficult to directly compare our work with prior art. Nevertheless, we use cyws~\cite{Sachdeva23} as a baseline as their formulation is the same as ours, except restricted to 2D transformations. To allow for a fair comparison, in addition to reporting the results using their open-sourced model, we also finetune their pretrained model on the same training dataset as ours for $100$ epochs (best model is picked using lowest loss on val set). We use the average precision metric to report our results, similarly to~\cite{Sachdeva23}.

\subsection{Training details}
\noindent We trained the model on 4$\times$ A40 GPUs for 50 epochs using the DDP strategy with a batch size of 16, where the best model is chosen using a validation set (COCO-Inpainted val set as in~\cite{Sachdeva23} + a KC-3D val set). Images were augmented with CropAndResize, HorizontalFlips and ColourJittering, and resized to $224\times 224$ due to the DINO-ViT~\cite{dino_vit} requirements. Since our training data is entirely synthetic, during training we use ground truth data for $\tau_{r\rightarrow q}(\cdot)$ rather than estimating it. The overall objective was optimised using Adam~\cite{adam} with learning rate of 0.0001 and weight decay of 0.0005.

\subsection{Results}
\noindent We evaluate our model, which is trained on synthetic data described in Sec.~\ref{sec:training_datasets}, on a diverse set of testing datasets as described in Sec.~\ref{sec:testing_datasets} with no further training/finetuning. Table~\ref{tab:results} contains the quantitative results in terms of average precision, while we show some qualitative predictions of our model in Figure~\ref{fig:predictions} and some failure cases in Figure~\ref{fig:failure}.\newline

\noindent \textbf{3D scenes} (KC-3D, RC-3D): From Table~\ref{tab:results} it is evident that our model produces impressive results on both synthetic and ``in the wild" real-world images. Particularly in the case of RC-3D, the model produces almost 4$\times$ better results than the baseline and performs well even in the most challenging setting when only RGB image pairs are available as input. Despite only having been trained on synthetic data, the remarkable performance of the model on real-world images validates the design choice of only using feature differences and not features directly (unlike cyws, where their co-attention module concatenates the cross-attended features). On the other hand, we observe a surprising result from the cyws model, which is able to produce impressive results on KC-3D dataset when finetuned on it. It is likely that the cyws model is ``over-fitting" to the Kubric setting given that the set of objects and scenes in Kubric~\cite{kubric} are limited (different scenes just have different random combinations from the same set of objects and backgrounds). Despite this fact, it is still interesting that the model is able to reason about changes despite not being 3D aware. Nevertheless, this performance does not generalise to real-world images as observed by its poor results on RC-3D.\newline

\noindent \textbf{2D/fixed-camera scenes} (Cyws test sets): In the 2D setting, we found that our model is often comparable but not strictly better than cyws. In particular, we observed that our model particularly struggles with detecting really small changes in comparison to cyws which is likely due to the fact that cyws operates at higher input resolution than us ($256\times 256$ vs $224\times 224$) and that the number of trainable parameters in our model is 31.5M which is much less than 49.5M in cyws. Furthermore, we found that a lot of the annotations (for COCO-Inpainted small and VIRAT-STD), are extremely small (handful of pixels, almost indiscernible to human-eye). This becomes problematic for our model when the input resolution is reduced to $224\times 224$. In addition, the ground-truth annotations for VIRAT-STD are noisy (both false positives and missing annotations, see Figure~\ref{fig:predictions} where our model predicts valid changes that are missing from ground truth).

\subsection{Limitations}
\noindent Despite the remarkable ability of our model to localise changes in real-world general two-view images, it suffers from a few limitations. A potential concern is the large size of the model. While the number of trainable parameters is 31.5M (which is less than 49.5M in cyws), accounting for the frozen DINO-ViT backbone~\cite{dino_vit} with 85.8M parameters, our total model size is roughly 117M parameters. This makes it much slower to train and infer from than cyws. However, it must be acknowledged that this large backbone comes with the added benefit of making our model generalisable to real-world images even though it has only been trained on synthetic data. Another potential cause of concern is that it relies on a good estimated registration, which in turn relies on reliable correspondences and depth, which may not be always available. On the other hand, it also means that as better correspondence extractors and monocular depth estimators become available, our model's results can improve at no additional cost.

\section{Conclusion}
\label{sec:conclusion}
\noindent In the ever-changing landscape of our world, the task of detecting changes in a 3D scene is daunting for both humans and machines alike. In this work we take a step closer towards solving this problem by automatically detecting changes in real-world images captured from significantly different viewpoints. Due to the lack of large-scale real-world training datasets for this problem, we propose a model that is trained entirely on synthetic data but can generalise to real-world scenes by leveraging the recent advances in 
Following previous works, we largely focused on detecting missing objects as they are easy to acquire and precise. However, we note that our model should not be confused with a ``(missing) object detector". While it is trained on object-centric datasets, our model can zero-shot detect all sorts of open-set changes.
\\[\baselineskip]
\noindent \textbf{Acknowledgements:} We would like to thank Jaesung Huh and Robert McCraith for their assistance with data collection, and Luke Melas-Kyriazi and Jaesung Huh for proof-reading the paper. This research is supported by EPSRC Programme Grant VisualAI EP/T028572/1 and a Royal Society Research Professorship RP\textbackslash R1\textbackslash191132.

{\small
\bibliographystyle{ieee_fullname}
\bibliography{shortstrings,vgg_local,egbib}
}

\end{document}